\documentclass[11pt]{article}

\usepackage[preprint]{acl}

\usepackage{times}
\usepackage{latexsym}
\usepackage{amsmath}
\usepackage[T1]{fontenc}

\usepackage[utf8]{inputenc}

\usepackage{microtype}

\usepackage{inconsolata}

\usepackage{graphicx}

\title{Repetitions are not all alike: distinct mechanisms sustain repetition in language models}

\author{First Author \\
  Affiliation / Address line 1 \\
  Affiliation / Address line 2 \\
  Affiliation / Address line 3 \\
  \texttt{email@domain} \\\And
  Second Author \\
  Affiliation / Address line 1 \\
  Affiliation / Address line 2 \\
  Affiliation / Address line 3 \\
  \texttt{email@domain} \\}

\author{
 \textbf{Matéo Mahaut\textsuperscript{1}},
 \textbf{Francesca Franzon\textsuperscript{1}},
\\
\\
 \textsuperscript{1}Universitat Pompeu Fabra,
\\
 \small{
   \textbf{Correspondence:} \href{mailto:mateo.mahaut@gmail.com}{mateo.mahaut@gmail.com}
 }
}

\begin{document}
\maketitle
\begin{abstract}
Large Language Models (LLMs) can sometimes degrade into repetitive loops, persistently generating identical word sequences. Because repetition is rare in natural human language, its frequent occurrence across diverse tasks and contexts in LLMs remains puzzling. 
Here we investigate whether behaviorally similar repetition patterns arise from distinct underlying mechanisms and how these mechanisms develop during model training. We contrast two conditions: repetitions elicited by natural text prompts with those induced by in-context learning (ICL) setups that explicitly require copying behavior. Our analyses reveal that ICL-induced repetition relies on a dedicated network of attention heads that progressively specialize over training, whereas naturally occurring repetition emerges early and lacks a defined circuitry. Attention inspection further shows that natural repetition focuses disproportionately on low-information tokens, suggesting a fallback behavior when relevant context cannot be retrieved. These results indicate that superficially similar repetition behaviors originate from qualitatively different internal processes, reflecting distinct modes of failure and adaptation in language models.

\end{abstract}

\section{Introduction}
Text output by Large Language Models (LLMs) diverges from human-generated text in systematic ways that occur consistently across models and tasks. One of the most striking examples of this divergence is repetition, where identical token sequences are persistently produced in cycles~\cite{fu2021theoretical}.  
It remains unclear why models generate repetitive patterns so commonly, and how this tendency develops over the course of training. 
Is it a learned strategy, or an innate tendency that emerges early and persists independently of training? 
Repetition has largely been treated as a unitary phenomenon, implicitly assuming a single underlying cause. Yet it can be elicited by diverse prompts and contexts, suggesting that distinct processes may contribute to sustaining repetitive behavior, as shown for other generation tasks~\citep{ortu2024competition}. 

We explore whether multiple mechanisms act concurrently in sustaining repetition, what specific roles they play, and how dedicated circuitry is organized to support them.
To this end, we investigate how repetition arises under different inputs, comparing two conditions with a controlled set of prompts: one in which repetition follows natural text, and another in which it is induced by the structure of the input through an in-context learning task.
Adopting a mechanistic interpretability approach, we analyze the model’s internal dynamics during repetition from three complementary perspectives: 
(1) developmental trajectory, tracking the emergence of repetitive behavior and attention-head activity across training (Sec. \ref{sec:dev}); (2) attentional focus, examining which tokens are attended during repetition (Sec. \ref{sec:focus}); and (3) confidence of next token prediction, analyzing output token probabilities (Sec. \ref{sec:confidence}).

Our analysis reveal that multiple, distinct mechanisms supporting repetition emerge at different stages of training and act jointly in the latest training steps. Specifically, repetition as an in-context learning behavior develops progressively, whereas an early-available, degenerate mechanism produces repetition when the model fails to attend to informative tokens. Over training, this early mechanism stabilizes, yet both mechanisms persist and interact to sustain repetitive behavior in fully trained models. Together, these findings suggest that the prevalence and persistence of repetition in LLMs arise from the interaction between these mechanisms.

\subsection{Related work}
Repetitions occurs across model sizes and architectures in response to various tasks and prompts~\citep{xi2021taming, shu2024analyzing, wang2024mitigating}, and share some fundamental properties, like a tendency to persist once initiated, especially when relying on top probabilities for next token selection, like with greedy decoding~\cite{holtzman2019curious, fu2021theoretical}. Prior studies explained its persistent as a self-reinforcing dynamic: once repetition begins, the model increasingly favors the same tokens when exposed to similar contexts~\citep{xu2022learning}. 
Repetition would thus emerge as a context-based strategy of copying previous tokens, akin to an in-context learning (ICL) ability~\citep{wang2025induction, yan2023understanding}, supported by specialized neurons and attention heads whose activation strengthens across cycles~\citep{hiraoka2024repetition, yao2025understanding}. 
Repetition may thus arise as an ICL skill, learned as a byproduct of optimization on some repetitive patterns already present in training data~\citep{li2023repetition}, which can influence the probability distributions the model learns from. Few details are known about how LMs would depart from regular text generation and start outputting repetitions~\cite{fu2021theoretical,yao2025understanding}; in fully trained models, repetition has been described as a fallback strategy arising under uncertainty, for example when lacking epistemic knowledge in fact-retrieval tasks~\citep{ivgi2024loops}. It is possible that training has effect on it, as it is more frequently present in smaller, shorter-trained models. Yet, even though properties of ongoing repetitions are known, such its self-reinforcing dynamics and context reliance, it remains unclear why and how models start develop and maintain repetition instead of converging toward more human-like generation.

\subsection{Motivations}
The aim of this study is to better understand repetitive behavior, by looking at how it develops during training, in different input contexts. Repetition can occur in response to diverse tasks and linguistic settings, likely reflecting multiple underlying causes. 
It may occur under uncertainty, when the model lacks the knowledge to select appropriate tokens, or paradoxically from overconfidence, when it assigns excessive probability to repeating sequences despite their degeneracy. Repetition may continue from already repetitive input, or unexpectedly arise after well-formed natural text.
This complexity suggests that repetition can arise and develop in different ways, as a response to diverse contexts. Here, we investigate how repetitive behaviors emerge during model training and how models develop mechanisms that sustain them. Specifically, we compare two conditions: one where repetition arises spontaneously following \textit{natural} well-formed sentences, and one prompting it as form of \emph{in-context learning}, where it is induced as a procedural copying behavior~\citep{wang2024mitigating, yan2023understanding, hiraoka2024repetition}.

\section{Experimental setup}
We conducted our experiments on Pythia 1.4B~\citep{biderman2023pythia}, which provides intermediate training checkpoints, enabling analysis of learning dynamics while remaining small enough to run extensive experiments on all checkpoints. Experiments are run on A30 GPUs. A run on all heads of a single layer, for a single checkpoint can take up to 10h. We ran experiments for all layers, on 10 different checkpoints.%

\textbf{Dataset.}
We designed a dataset to contrast two types of prompts, both eliciting repetitive outputs.
First, we built a \textbf{raw set} of prompts by randomly sampling 1,000 well-formed human sentences from a Minipile partition of the Pile~\citep{kaddour2023minipile, gao2020pile} and truncating each to 32 tokens. These sequences were then input to the model, and continuations up to 1,000 tokens were generated using greedy decoding, a method known to induce repetition~\citep{holtzman2019curious, li2023repetition, xu2022learning}.

From the raw set, we retained as \textbf{Natural} prompts only those sequences that spontaneously trigger repetitions under greedy decoding: specifically, cycles of more than one token that persisted until the end of generation (excluding repetitions of the end-of-text token). If a cycle did not start immediately after the original 32-token extract, we included in the prompt  the model’s output up to the first token in the cycle. Constructed this way, when a prompt from this dataset is submitted to the model, it outputs cycle from the first generated token.

For the \textbf{ICL} prompts, we sampled input from the set that did \textit{not} trigger repetitions at all during the 1000 token generation. By taking those raw sequences and repeating  them at least once in the prompt, we create prompts which induce repetition in the model.

\textbf{Repetition cycles}. We formally define the notion of repetition cycles as used in this work.
A sequence of tokens \( \{w_t\} \) is repetitive with cycles of length \( n \) if there exists an index \( T \) such that for all \( t \geq T \):
\[
w_t = w_{t+n}, \quad \forall t \geq T
\]

where \( \{w_t\} \) is the sequence of tokens. \( T \) is the starting point of periodicity. \( n \) is the \textit{cycle length}. This means that for \( t < T \), the sequence may be arbitrary, but from \( T \) onward, it follows a periodic pattern. 

\textbf{Cycle numbering.}  In what follows, we indicate the first repetition of a cycle as \emph{cycle 1}. By \emph{cycle 0}, we indicate the input sequence not including the tokens repeated by the model. In the natural prompt case, this is solely the prompt. In the ICL case, cycle 0 is an empty prompt, which we do not plot. Cycle 1 is a single iteration of the sequence, insufficient to expect a repetitive generation. From cycle 2, the model has a pattern to detect and could start repeating.

\section{Development}
\label{sec:dev}
\subsection{Behavioral path}
We examine the developmental trajectory of repetition in both Natural and ICL prompts. Specifically, we test whether the ability to repeat in the ICL condition emerges progressively over training, as observed for other in-context learning abilities \citep{olsson2022context}, and whether it develops across training or is early present as a default in natural setting. Using the same raw set of 1,000 prompts from the Pile, we quantified the proportion of outputs exhibiting repetition under each condition. We then tracked whether the same prompts remained in the repeating category across checkpoints to assess the stability and development of repetition over time.

\begin{figure*}[t]
    \centering
    \includegraphics[width=0.95\linewidth]{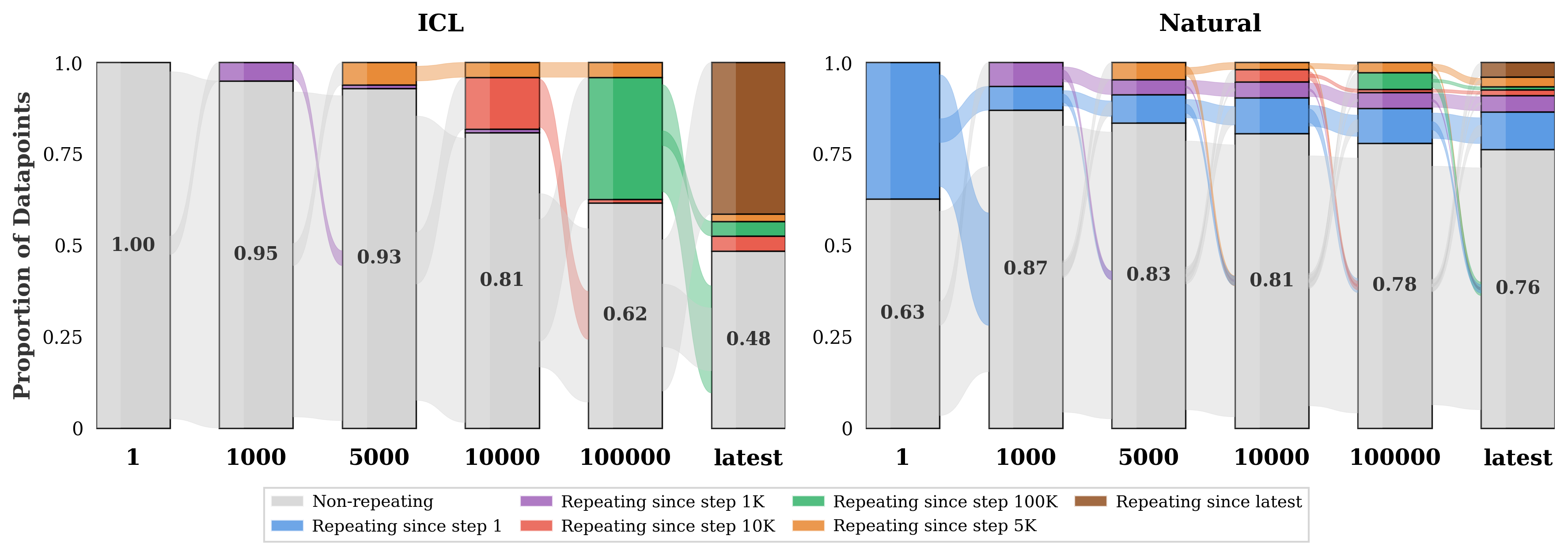}
    \caption{Proportion of repetitions across checkpoints in the ICL (left) and natural (right) settings. At each training step, the same prompts are re-input to the model. Flows indicate whether they remain in the repeating (colored) or non-repeating (light grey) category across steps. Early-emerging repetitions are largely preserved across checkpoints in the natural setting while ICL is late emerging.}
    \label{fig:alluvial_proportions}
\end{figure*}

Figure~\ref{fig:alluvial_proportions} illustrates the results: in the natural setting, repetitions emerge as an \textit{early-available} behavior starting from the first training timestep, with initially present repeating sequences still significantly represented at the latest training step. Once the model is fully trained, natural repetition encompasses half sentences that were repeating since the initial training stage, and half that were acquired progressively, throughout training. 

By contrast, in the ICL setting, no repetitions appeared at the first timestep; the proportion of repetitions increased gradually, with prompts moving in and out of the repeating category, suggesting that repetition develops progressively as an \textit{acquired} behavior.
The increasing proportion suggests that the model learns to perform ICL repetition more effectively and generalizes it to new inputs with the appropriate contextual structure.

\subsection{Attention head activations}
\label{sec:attention}
Having established distinct developmental trajectories for natural and ICL repetition, we next assess whether these behaviors rely on different circuitry. Specifically, we ask whether the developing ability to match context recruits a network specializing over time, and whether a circuit is already present to sustain the early-emerging repetition observed in the natural setting.
We focus in particular on attention heads, which are known to support pattern-matching operations such as in-context repetition~\citep{olsson2022context, crosbie2024induction, zheng2024attention}: we compare their activity across checkpoints in the two conditions.

\subsubsection{Method.} To understand the contribution of each attention head to repetitions, we map each head to the model's output space using an affine transformation, and the model's unembedding matrix~\citep{belrose2023eliciting}. 
We assess which output token can be predicted from the attention-head probability distribution, focusing on the first four cycles generated after each prompt.

If $k$ is the number of tokens of a sequence, and each layer $l$ has $n$ attention heads with input and output dimension of $d$, then the update made to the residual stream by an attention head $a$ can be noted:
\[
A^{l,n} = a^{l,n} \left(\text{norm}(x^{l-1}) \right)
\]
with $x \in {R}^{d \times k}$ the input from the residual stream, and $norm$ layerwise normalization. We train an affine transformation to map each attention head's output to the unembedding space:
\[
Z^{l,n} = W^{l}_v A^{l,n} + b_{\text{vocab}}
\]
with \( W_v \in {R}^{d \times d} \) the \textit{trainable weight matrix} mapping the hidden representation to the unembedding space and \( b_{\text{vocab}} \in {R}^{d} \) the \textit{bias term} added to each logit.

We use the contrastive method from~\citet{ortu2024competition} to highlight the contribution of each head to the probability of outputting the next token in a given cycle, rather than the next most likely, non-cyclical token.
In practice, we subtract the logits of cycle token $ctok$ from the most likely token that is not $ctok$. This next most likely token, $ntok$, corresponds to the token most likely to interrupt the cycle: \[
contrast=softmax(Z^{l,n}W_u^{(ctok)} - Z^{l,n}W_u^{(ntok)})
\]
with $W_u \in {R}^{d \times V}$ the unembedding matrix and $V$ the vocabulary size.

Positive values of this contrast mean the attention head favors the token that continues the cycle, and negative values mean that it favors another token, that is not in the cycle. 

\subsubsection{Training.} As in~\citet{belrose2023eliciting}, we find the parameters of the affine transformation using stochastic gradient descent to minimize Kullback-Leibler divergence between an attention head's distribution and the distribution of the final layer, right before the unembedding matrix. We perform this training on a set of 10k prompts extracted from minipile~\citep{gao2020pile,kaddour2023minipile}, which we later discard from further experiment to avoid data contamination. We use the representation extracted for the final $k$th token to represent the entire sequence.

\begin{figure*}
    \centering
    \includegraphics[width=1\linewidth]{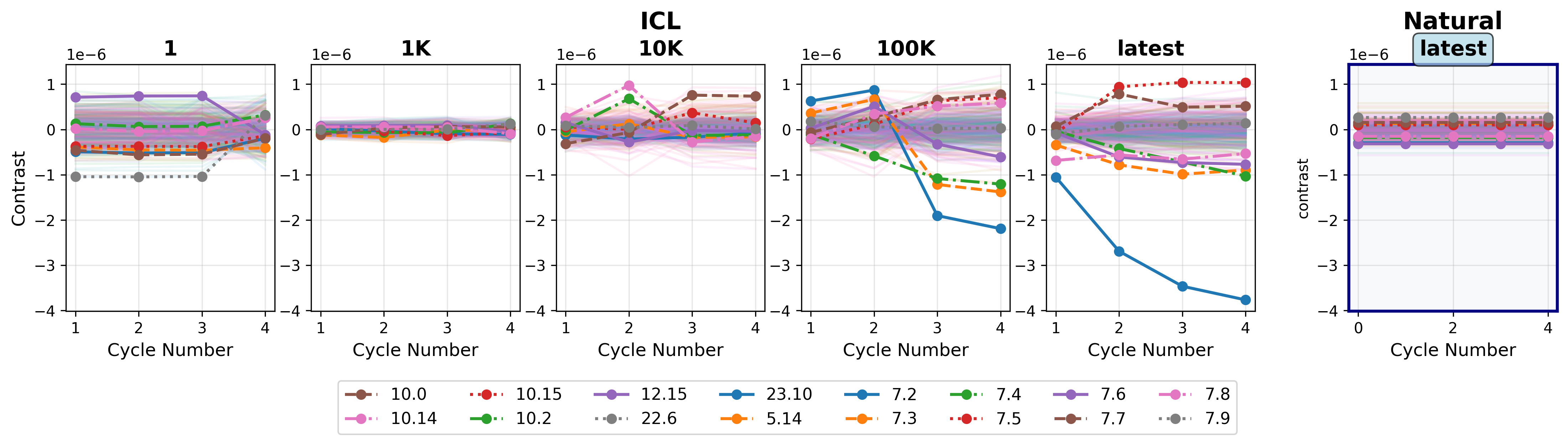}
    \caption{Attention head contrast between outputing a repeating token--above 0--or non repeating--below 0. On the left are multiple steps of ICL training, showing progressive specialisation of specific heads in either directions. On the right is the final training step for the natural dataset. Heads do not contribute either way. Heads with biggest variation across cycle number are put in full opacity. In the legend, we use the layer.head-number format.}
    \label{fig:evolution_attention}
\end{figure*}

\subsection{Results}
\label{sec:attention-results}
Fig.~\ref{fig:evolution_attention} shows the evolution of each attention head's contribution to the repetitive behaviour. In the ICL setting, contrast values show that repetition is characterized by the activation of a few heads, which specialize throughout training. 
Repetition in the natural setting does not rely on the same circuits, and is characterized by near 0 contrast on all attention heads at all intermediary step. This remains true until the last step (we graphically illustrate this last one in Fig. \ref{fig:evolution_attention}). 
The results suggest that ICL-induced repetition engages a distinct and progressively specialized circuit of attention heads, unlike repetition from natural text, which operates independently of this system. We next detail how this circuit develops.

\textbf{ICL before step 10k:} As observed in the previous section (see Fig~\ref{fig:alluvial_proportions}, left panel), the proportion of ICL-based repetitions is minimal until step 10k. Expectedly, as we observe close to no ICL repetition in this first phase of training, we find that attention heads have random contrast values from cycle to cycle, centered around 0. In this early phase, the model has not learned yet to perform ICL-based repetition. 
 
\textbf{ICL at steps 10k/100k:} The majority of heads have very low contrast values, between -.5e-6 and .5e-6. There is a general shift observed between cycles 2 and 3, when the sequence has been repeated and it becomes possible to detect it. On cycle 3, we observe that a majority of heads change contrast value rapidly. 
Heads which showed a positive contrast value, in favor of repetition, will suddenly drop well below zero, and vice versa. We understand this change to mean that the heads have detected the pattern, and will either act to contrast it or encourage it. Before, they could only focus on elements of context, it was impossible to detect a repetitive structure. 
We take this to be characteristic of  an intermediary   repetition behavior during the training of the LLM. It uses specialized heads, but they emerge only once the repetitive pattern is completely available and sufficiently long to create a stable context. 

\textbf{ICL at latest training step:} As previously, only a few specialized heads contribute to repetition, while all others remain have very low activations. The strongly activated heads (heads 2 to 8 from layer 7) have increasingly polarized values as the number of cycles increases. At the latest training step, those specialized head's contrast corresponds to results from the literature reporting increasing probabilities for repetition as number of cycles increases. At this point, there is no radical change after the cycle has been repeated twice, specialized heads support repetition before without relying on a previous repetitive pattern: this pattern agnostic evolution can explain the increased proportion of repetitions elicited at this step (Fig.~\ref{fig:alluvial_proportions}).

\textbf{Natural at latest training step:} No activation of attention heads beyond the noisy -.5e-6 and .5e-6 thresholds is measured for natural. When the number of cycles increases, there is no change in contrast value for any given head. %
The absence of contrast change with the evolution of the context when cycle number increases will be further analyzed in the next section, which investigates the tokens the different heads focus on.
    
Similarly, as shown in Appendix \ref{sec:mlp} MLP contribution to repetition is much stronger for induced repetition than for the natural ones. In both cases, only the later layers contribute, and the contribution of MLP layers is not detected by the probes during the first phases of training.

\section{Attention head focus}
\label{sec:focus}
While ICL-induced repetition depends on dedicated attention heads, the natural setting shows no clear head specialization yet still produces repetition. 
This suggests that repetition may arise either from alternative information sources or from a lack of informative context, which could prevent a meaninful continuation. We examine this possibility by analyzing the tokens each head attends to during generation.
We therefore focus on the linguistic information present in the prompt context that attention mechanisms are expected to capture to generate coherent continuations. 
To this end, we analyze the attention weights assigned to different classes of input tokens across the two conditions.

\subsection{Method} 
\begin{figure}
    \centering
    \includegraphics[width=0.95\linewidth]{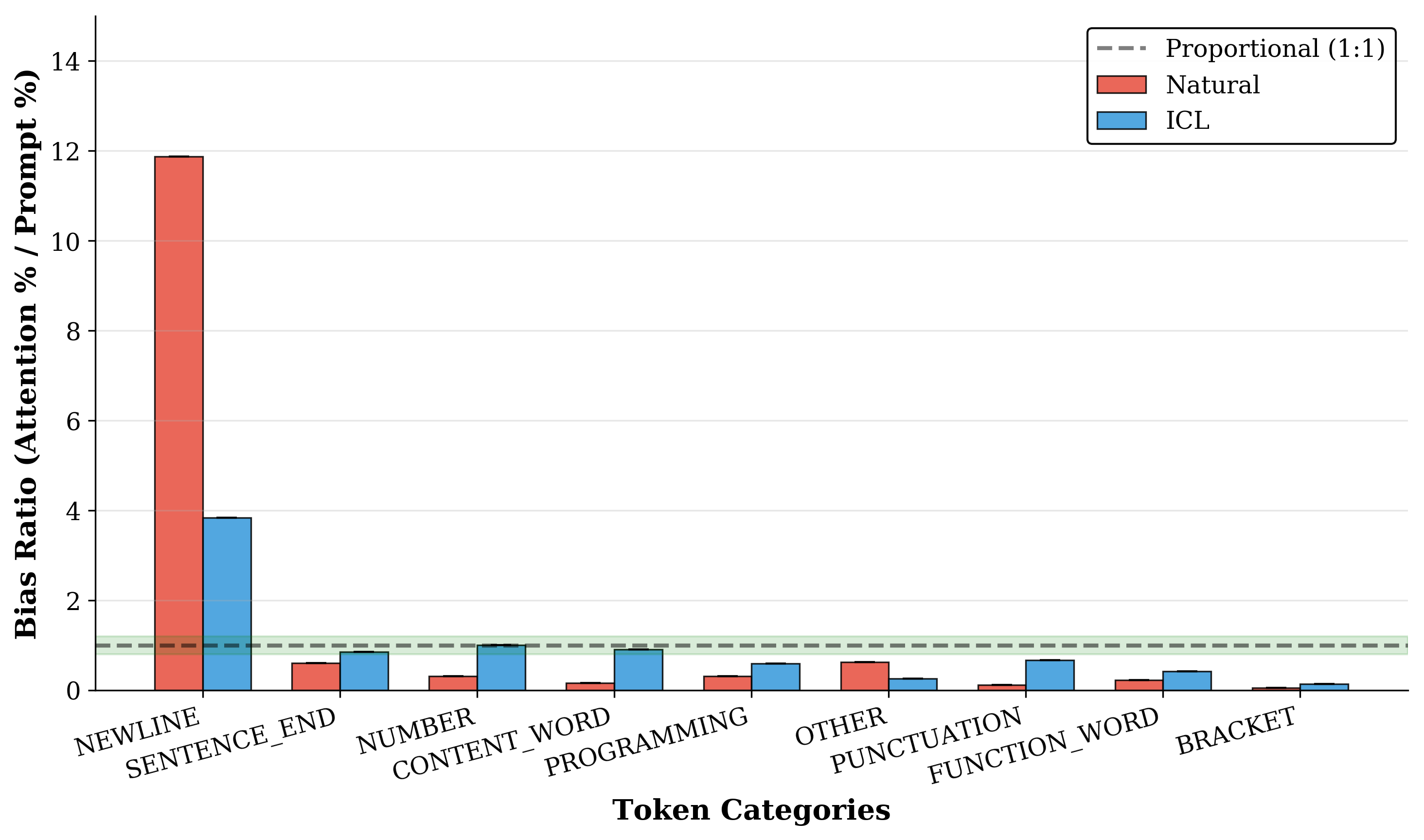}
    \caption{Average attention head focus on different token categories for the two different datasets. Attention focus is reported as a ratio of the relative importance given to tokens compared to their proportion in the original dataset. The leftmost bar on the left indicates that on average, attention heads will assign 9.9 times more attention it would if attention was equally distributed between all tokens in the dataset.}
    \label{fig:attention_focus}
\end{figure}

Words from the prompts are categorized according to their linguistic function, depending on the different types of information they might be carrying:
1)  Content Words, carrying relevant semantic information (such as \emph{nouns, adjectives}, and \emph{verbs}), which occur most frequently in our prompt set, are classified as .
2) Structural Words and tokens which lack semantic content. We segment those into different categories (e.g. as seen in Fig. \ref{fig:attention_focus}, \emph{newline, bracket, sentence-ending punctuation, other punctuation}), and Other formatting tokens.

An attention head transfers information from previous tokens in a sequence to the current token: each token gets a specific attention weight, corresponding to it's relevance in the given context. %
In Fig. \ref{fig:attention_focus} we average the proportion of attention weights given to each category of words in the sequence over all attention heads in the model. To control for differences due to the frequency of each category of words in the dataset, we report a ratio--the proportion of attention given to a given category divided by the proportion of words of that category in the dataset. Here, if all categories had a bias of 1 (represented by a dotted line in Fig. \ref{fig:attention_focus}) we would have a uniform distribution of the attention weights where all words in every prompt would be attended equally.

\subsection{Results}
Attention in the two prompting conditions focuses on different types of tokens. In the natural condition, compared with ICL, it is predominantly directed toward tokens carrying little semantic or linguistic content (most notably \textit{newline} tokens) and less toward semantically informative tokens such as content words.

Tokens forming a \textit{newline} are very attended on average by all heads in the model (We provide a breakthrough of newline and content word attention focus at every layer in Appendix \ref{sec:attention}). This pattern is notable also in the ICL case, with \textit{newline} given 3.1 times more weight than expected from the uniform distribution. For Natural prompts, the distribution of attention weights is even more biased towards \textit{newline}, with a ratio of 9.9. 

We emphasize that while Content words are within a standard deviation of the uniform distribution in the case of ICL, a majority of words in our prompts are content words (This can be visualised in Appendix \ref{sec:appendix}, Fig. \ref{fig:layer4_full} where the distribution of each dataset in the different categories is explicited. Attention distribution for layer 4 is also provided as an example) Content words are where most semantics will be found. Attending content words is an expected behavior in most contexts for LLMs when performing next token prediction. They are nonetheless vastly ignored when we input prompts from the Natural dataset, and get on average less than half of the attention we could expect with a uniform distribution.

We make comprehensive causal tests to determine whether it is the \textit{newline} tokens in themselves that cause repetition (adding newlines, removing newlines, forcing attention heads to give additional / less weight to newline). All these experiments lead to 0 changes on our experimental subset of 100 prompts of each dataset.

\begin{figure}
    \centering
    \includegraphics[width=0.95\linewidth]{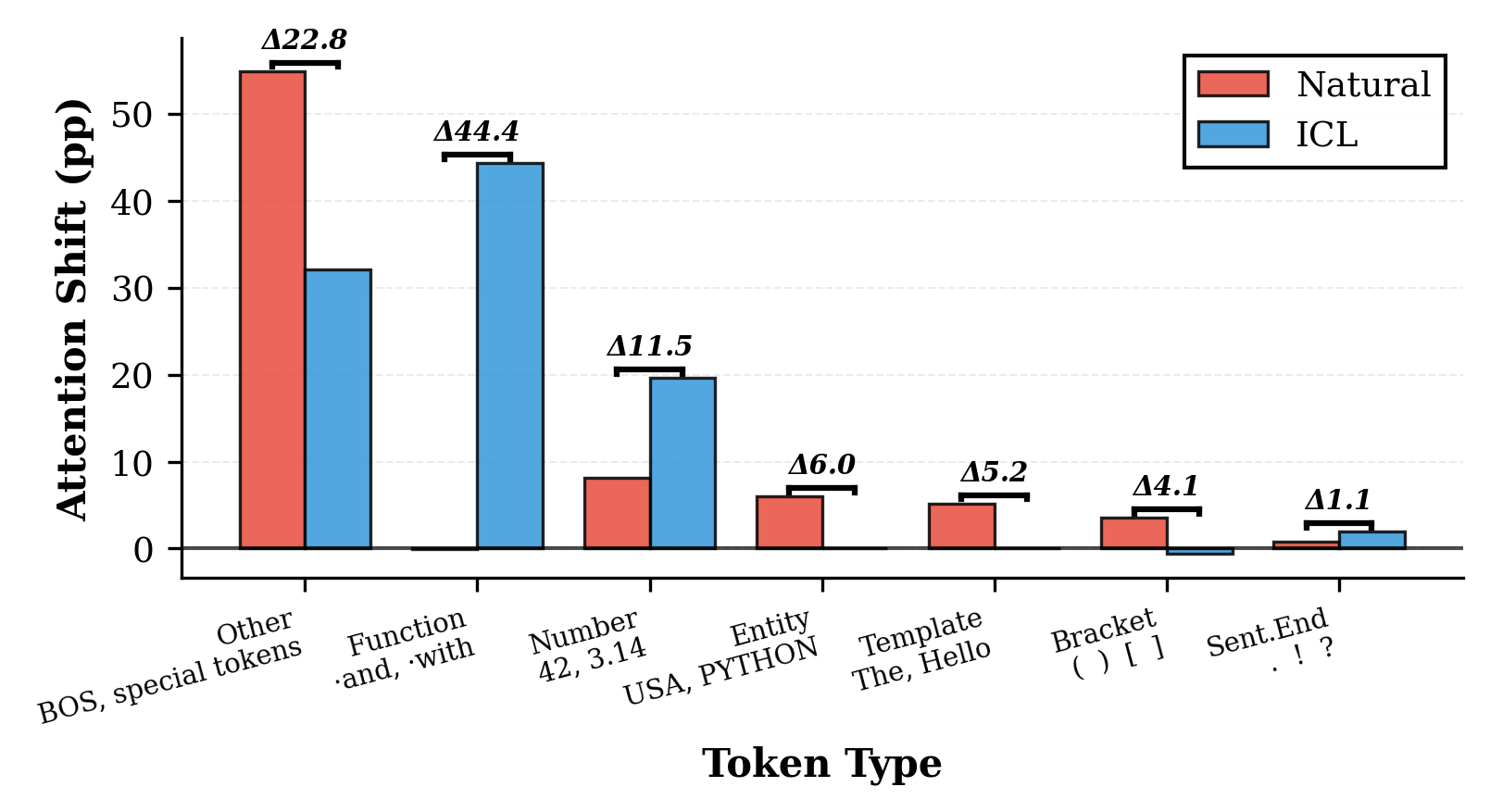}
    \caption{Percentage of attention weight that shifts to different token categories when we remove all\textit{ newline }tokens from the sequence. While ICL fallbacks on function words primarily, in natural sequences attention heads mostly focus on structural tokens, specific to the tokenizer.}
    \label{fig:fallback}
\end{figure}

A high bias to attend \textit{newlines} and a bias to avoid Content words do not \textit{cause} repetition from the natural prompt dataset, but they nonetheless inform us that repetitions co-occurs with attention heads avoiding semantic information; \textit{newline}, like other structural tokens has previously been observed to be completely replaceable, without any impact on next token prediction \cite{rakotonirina2024evil}. Attention heads massively focusing on such a token helps explain what we observed in Section \ref{sec:attention-results}: the attention heads are not contributing significantly to repetitive behavior as they are looking at overall uninformative tokens. We further confirm this disregard for semantics in Fig. \ref{fig:fallback} by observing which words receive attention weight when we remove all\textit{ newline }tokens from the prompts. Attention from the\textit{ newline }tokens mostly shifts to other tokens that do not hold semantic information, such as structural tokens specific to tokenization or brackets. In induced repetition on the other hand, the attention weight is transferred to function words (e.g. "and", "with", etc.) or numbers when they are available. This further highlights that sequences that spontaneously lead to repetition are those where attention strongly disregards content, and mostly focuses on structural, mostly semantic-less tokens.

\section{Confidence}
\label{sec:confidence}
We finally assess whether the two repetition mechanisms yield different levels of confidence in the fully trained model. Specifically, we ask whether the learned ICL strategy and the early-arising natural repetition reflect distinct confidence profiles, as a hallmark of their different generation dynamics.

Previous studies have shown that models tend to repeat under uncertainty~\citep{ivgi2024loops}, but also that repetition can arise from excessive reliance on contextual cues~\citep{fu2021theoretical,xu2022learning}. 
To evaluate the model’s confidence during repetition, we compute the entropy of the next-token distribution, derived from the logits over the first four cycles of 1000-token greedy generations. High-entropy distributions indicate lower confidence, as probability mass is spread across many tokens, whereas low entropy reflects peaked, high-confidence predictions.

\begin{figure}
    \centering
    \includegraphics[width=0.85\linewidth]{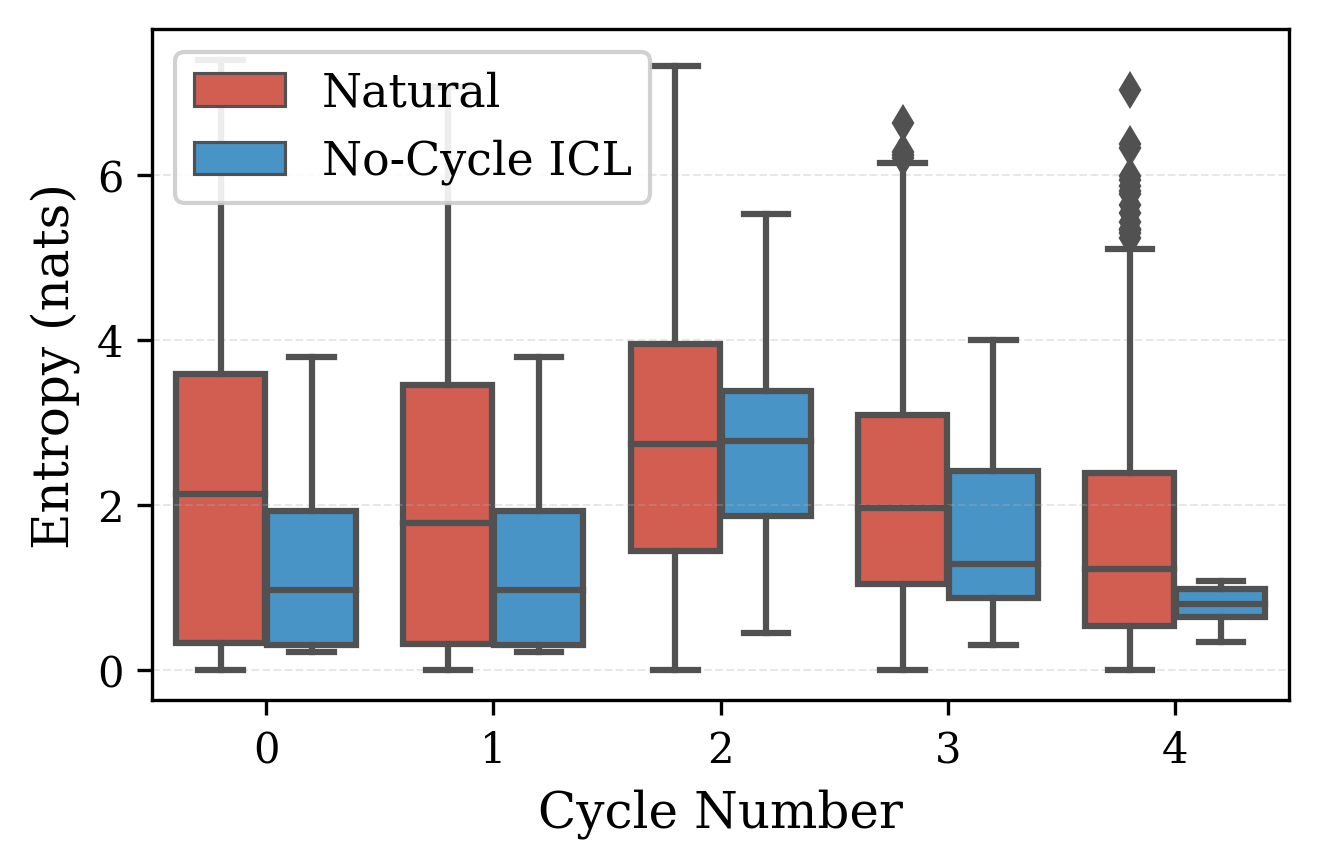}
    \caption{Entropy of the token probability space for the first token in a cycle, for different number of repeated cycles. Low entropy is when probability mass from the distribution is concentrated in a few tokens. Higher entropy would spread probability mass to more tokens. }
    \label{fig:entropy}
\end{figure}

The results, plotted in Fig.~\ref{fig:entropy} show that the distribution between quartiles is much wider in the natural case, with more outlier values. As the number of cycles increases, in both cases both datasets see their median entropy decrease, with a slightly faster decrease for ICL. Within-dataset differences are also present: while all ICL datapoints end up with the same low entropy with 99\% of data concentrated between 0 and 1 in cycle 4, sentences from the natural dataset repeat with more uncertainty, with some outlier sentences even keeping an entropy above 6 at cycle 4, effectively increasing entropy slightly. 
This suggests that natural repetitions start with lower confidence, consistent with the view that, in the fully trained model, ICL repetition represents a consolidated strategy reinforced after the second cycle, whereas natural repetition arises from incomplete information retrieval.
Taken together, these findings put together previous accounts, suggesting that both uncertainty and overreliance on context can lead to repetition, each reflecting concurring mechanisms in repetition generation.

\section{Discussion and Conclusions}
This work investigates whether behaviorally similar patterns of repetition in language models arise from distinct underlying mechanisms, and how these mechanisms evolve during training to jointly produce repetitive behavior. We find that repetitions elicited by natural language and by in-context learning (ICL) prompts display distinct signatures across multiple levels: behavioral development, underlying circuitry involving attention heads and MLPs, attention allocation, and model confidence.

Our analyses show that ICL-induced repetition emerges as a \textit{learned} strategy based in the model’s ability to detect and exploit context structure, becoming increasingly refined across checkpoints. A dedicated circuit of attention heads, supported by late-layer MLP specialization, develops to sustain this ability. The presence of stable input patterns enhances model confidence in token selection, consistent with previous findings: longer repeating contexts, encompassing more cycles, correlate with lower entropy and higher probabilities for repetitive tokens. This suggests that even when repetition originates naturally, the learned ICL mechanism can take over, reinforcing and sustaining repetitive output.

In contrast, natural repetition appears very early in training; although some of these early repetitive patterns are later unlearned, others persist through to the final model. This suggests that not all forms of repetition are learned—some may instead stem from architectural or data-driven biases that manifest from the outset. 
Repetition in this condition lacks a clearly identifiable circuit, indicating that the model does not rely on a specific procedure. This pattern is consistent with the uncertainty-driven behavior described by \citet{ivgi2024loops}.
Qualitative inspection of attention weights reveals a disproportionate focus on tokens with little semantic content—such as punctuation or syntactic markers, while largely ignoring semantically informative words. These tokens convey limited information, and models cannot extract meaningful content from them~\citep{kervadec2023unnatural,rakotonirina2024evil}, leading the language model to fall into a degenerate state.
Because this behavior resembles previously observed fallback patterns in attention, and given the absence of a causal link between structural tokens and repetition, we interpret natural repetition as a fallback behavior that emerges when the model fails to retrieve or attend to relevant information through either attention or MLP pathways. The causal mechanisms leading into this and their practical implementation require further investigation.
Understanding the origins of these early-emerging patterns remains an open and compelling direction for future work.

\subsubsection*{Acknowledgments}
Our work was funded by the European Research Council (ERC) under the European Union’s Horizon 2020 research and innovation programme (grant agreement No. 101019291). This paper reflects the authors’ view only, and the ERC is not responsible for any use that may be made of the information it contains. The authors also received financial support from the Catalan government (AGAUR grant SGR 2021 00470). 
We are grateful to Marco Baroni, Germán Kruszewski,  Iuri Macocco, Beatrix Nielsen, Carraz Rakotnirina for their valuable discussions and constructive feedback . We also thank Alberto Cazzaniga, Diego Doimo, and Francesco Ortu, for generously sharing details of their implementation and providing clarifications on their work. LLM assistance was used for figure cosmetics and some reformulating.

\section*{Limitations}
Our study focused on Pythia, as its checkpoints are freely accessible. We chose a relatively small-scale model as it is more prone to repetition~\citep{ivgi2024loops} and moreover, its small size allowed for intensive experimentation while avoiding intense computation costs. Future work should extend this analysis across models of varying sizes and architectures to confirm these results as a fundamental property of LLMs. %
\section*{Risk}
While repetition itself is a degenerate but harmless form of failure, understanding its mechanisms provides valuable insight into the internal dynamics of LLMs. By identifying how such non-functional behaviors arise, we can better characterize the pathways that may also produce more harmful outcomes, such as biased or toxic text generation.

\bibliography{custom}
\newpage
\appendix

\section{MLP contribution to repetition}
\label{sec:mlp}
Like attention, the MLP is a main component of LLMs. Previous works have studied it's contribution to different behaviors, including repetition\citep{hiraoka2024repetition}. We apply the method from Section \ref{sec:attention} but this time on each MLP module in the model. For each of them, we get a contrast value corresponding to how much the next token in a repeating token is more likely than the next most likely non-repeating token.  

Fig. \ref{fig:layer_bias} shows that unlike with attention, none of the MLP layers go against repetition. Contribution of the MLP remain weak for natural prompts, while they are very strong in the ICL setup, reaching up to 0.5, many orders of magnitude above observed contributions for the attention heads. In both cases, the first non-zero contrast reported is on step 10 000 of training. For ICL this corresponds to results observed in Section \ref{sec:dev}. For Natural, absence of MLP signal is additional evidence towards a default architecture or dataset behavior.

\begin{figure*}[b]
    \centering
    \includegraphics[width=1\linewidth]{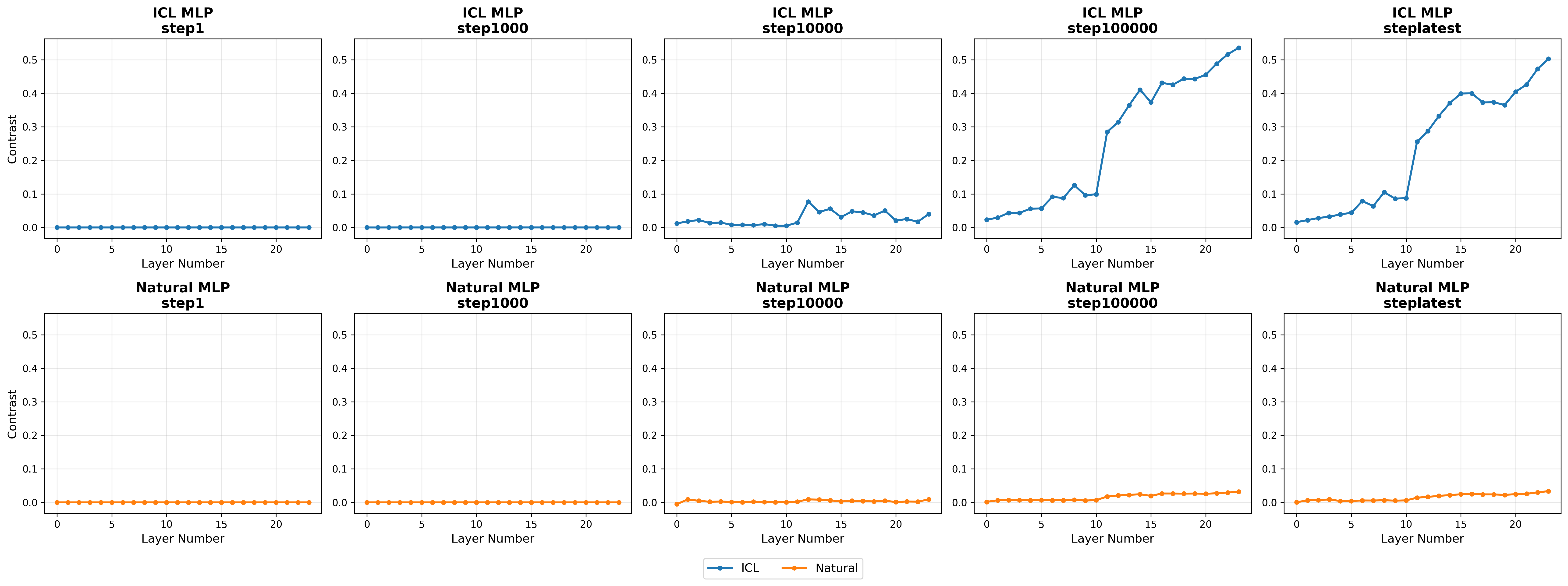}
    \caption{Contribution of each layer to promoting the repetitive token rather than any other token for both ICL, in blue in the top row, and Natural prompts, orange in the bottom row. Columns show the emergence of an effect during training at different training steps.}
    \label{fig:mlp}
\end{figure*}

\section{Attention focus across layers}
\label{sec:appendix}
\begin{figure*}[b]
    \centering
    \includegraphics[width=1\linewidth]{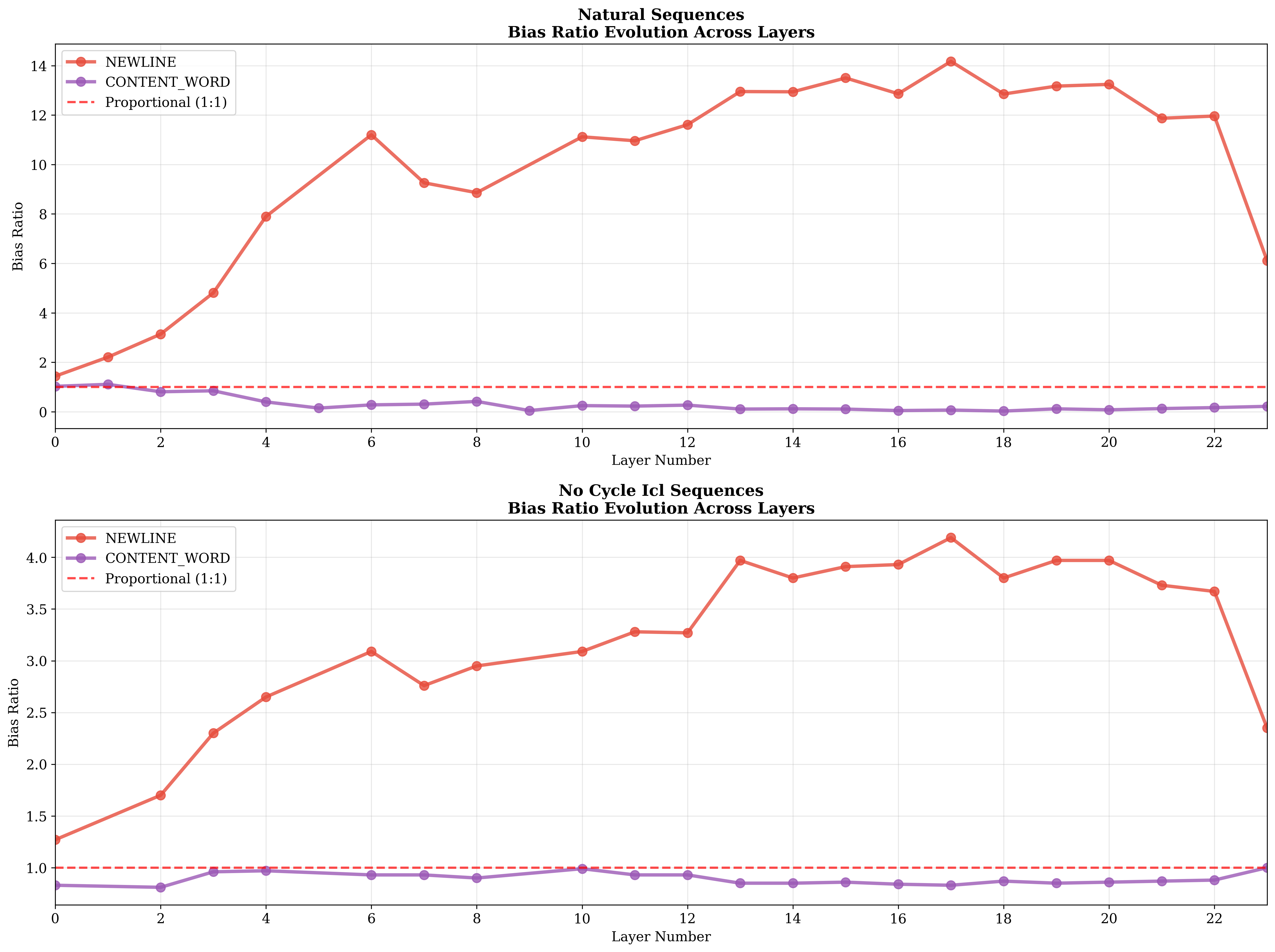}
    \caption{In each layer, the average proportion of attention weight given to\textit{ newline }token (red) or any content word (blue). Results are provided relative to how many times those tokens appear on average in the input. The dotted line represents a baseline distribution where the attention head would attend all words in the input sequence equally.}
    \label{fig:layer_bias}
\end{figure*}

Following methods described in Section \ref{sec:focus}, Figure \ref{fig:layer_bias} look at the ratio of attention weights focusing on either\textit{ newline }tokens, or any content words, in both settings. We observe that bias in favor of the newline token follows very similar patterns of relative growth. Bias increases throughout the first half of the network, and then hits a ceiling around layer 13. A drastic drop is observed in the final layer. These variations do not co-occur with the sudden MLP activation (Fig. \ref{fig:mlp}) in layer 10.   

\begin{figure*}
    \centering
    \includegraphics[width=1\linewidth]{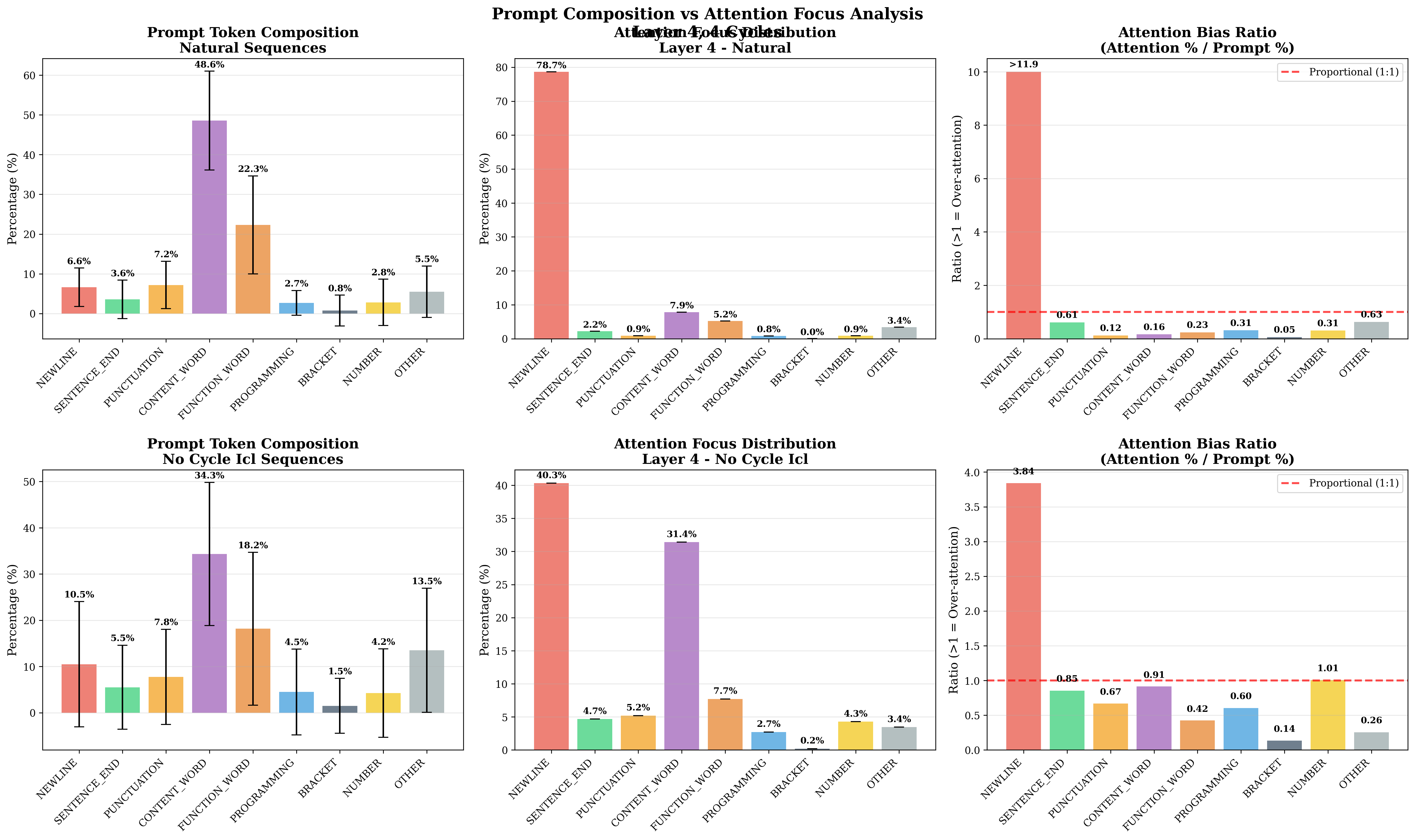}
    \caption{Complete analysis for layer 4, showing in the first column the percentages of different word types accross the datasets, with error bars showind standard deviation across prompts. Plots in the second column show proportion of attention assigned to each word category. Finally the third column is the ration of attention to a given category by the proportion of words in that category for the dataset. The top line gives results for Natural prompts while the second line gives results for the ICL prompts.}
    \label{fig:layer4_full}
\end{figure*}

Fig. \ref{fig:layer4_full} is a complete overview for attention head analysis in the fourth layer, allowing a more detailed comparison of the proportion of each category of word in each dataset, and the relative importance of content words.

\end{document}